\documentclass[letterpaper]{article}
\usepackage{aaai}
\usepackage{times}
\usepackage{helvet}
\usepackage{courier}
\usepackage{graphicx}
\usepackage{lipsum}
\usepackage{hyperref}
\begin{document} 
\title{Behavioral Cloning via Search in Video PreTraining Latent Space}

\author{Federico Malato*\\
University of Eastern Finland\\
fmalato@uef.fi\\
\And 
Florian Leopold*\\
University of Bielefeld\\
fleopold@techfak.uni-bielefeld.de
\And
Amogh Raut\\
Indian Institute of Technology BHU\\
\AND
Ville Hautam{\"a}ki\\
University of Eastern Finland\\
\And
Andrew Melnik\\
University of Bielefeld\\
}

\maketitle

\begin{abstract}
\begin{quote}
Our aim is to build autonomous agents that can solve tasks in environments like Minecraft. To do so, we used an imitation learning-based approach. We formulate our control problem as a search problem over a dataset of experts' demonstrations, where the agent copies actions from a similar demonstration trajectory of image-action pairs. We perform a proximity search over the BASALT MineRL-dataset in the latent representation of a Video PreTraining model. The agent copies the actions from the expert trajectory as long as the distance between the state representations of the agent and the selected expert trajectory from the dataset do not diverge. Then the proximity search is repeated. Our approach can effectively recover meaningful demonstration trajectories and show human-like behavior of an agent in the Minecraft environment.

\end{quote}
\end{abstract}

\section{Introduction}
This study was motivated by the MineRL BASALT 2022 challenge \cite{ShahDragan2021}. In the challenge, an agent must solve the following tasks: find a cave, catch a pet, build a village house, and make a waterfall \cite{ShahDragan2021}. The provided dataset of experts' demonstrations contains trajectories of image-action pairs. Additionally, both the MineRL BASALT dataset and environments do not contain reward information. Therefore, our primary focus was on Behavioural Cloning (BC) and Planning \cite{beohar2022planning}\cite{beohar2022solving} methods to address the tasks, rather than deep reinforcement learning (DRL) \cite{bach2020learn}\cite{schilling2021decentralized}.

\section{Methods}

A dataset of expert demonstrations solving the following tasks was provided \cite{ShahDragan2021}: find a cave, catch a pet, build a village house, and make a waterfall. Each episode is a trajectory of image-action pairs. No reward information is provided. 

In our approach, we use experts' demonstrations to reshape the control problem as a search problem over a latent space of partial trajectories (called \textit{situations}). Our work assumes that:
\begin{itemize}
\item Similar \textit{situations} require similar solutions or actions.
\item A \textit{situation} can be represented in a latent space.
\item The \textit{situations} latent space is a metric space. Therefore, we can assess the numerical similarity between any two \textit{situations}.
\end{itemize}

\begin{figure}[!t]
    \centering
    \includegraphics[width=\columnwidth]{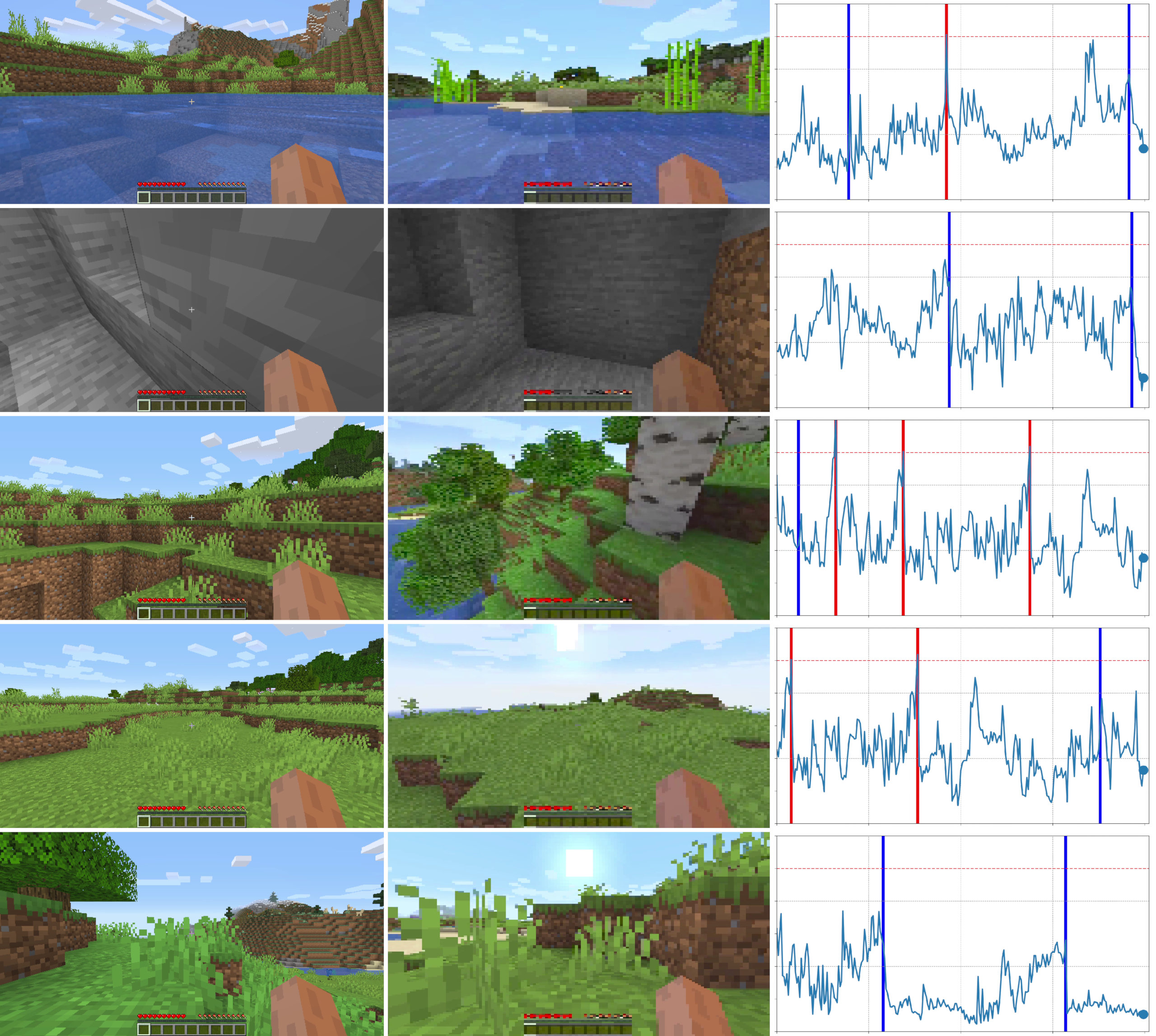}
    \caption{Similarity and divergence of five pairs of simulator and dataset \textit{situation} trajectories.
    \textbf{Left column}: current frame from the MineRL environment. \textbf{Middle column}: current reference frame from the dataset trajectory that the agent follows. \textbf{Right column}: L1-distance plot between VPT embedding points of simulator and dataset \textit{situations}. The current step shown as RGB images in the \textbf{Left} and \textbf{Middle} columns is highlighted by the rightmost bold dot marker in the \textbf{Right} column. X-axis indicates 256 time steps. Y-axis indicates L1 distance between two embedding points in the VPT latent space and ranges from 0.1 to 0.4. The deviation threshold is show by dashed red horizontal lines L1 = 0.35. Colored vertical lines mark new search events to find the most similar situation in the dataset. Blue lines indicate time-based initiated searches, red lines indicate deviation-threshold based initiated searches.  Gray dotted lines: every 64 steps in the X-axis and every 0.1 steps in the Y-axis.}
    \label{fig:divergence}
\end{figure}

\subsubsection{Video PreTraining (VPT) model}
\begin{figure}[!t]
    \centering
    \includegraphics[width=2.5in]{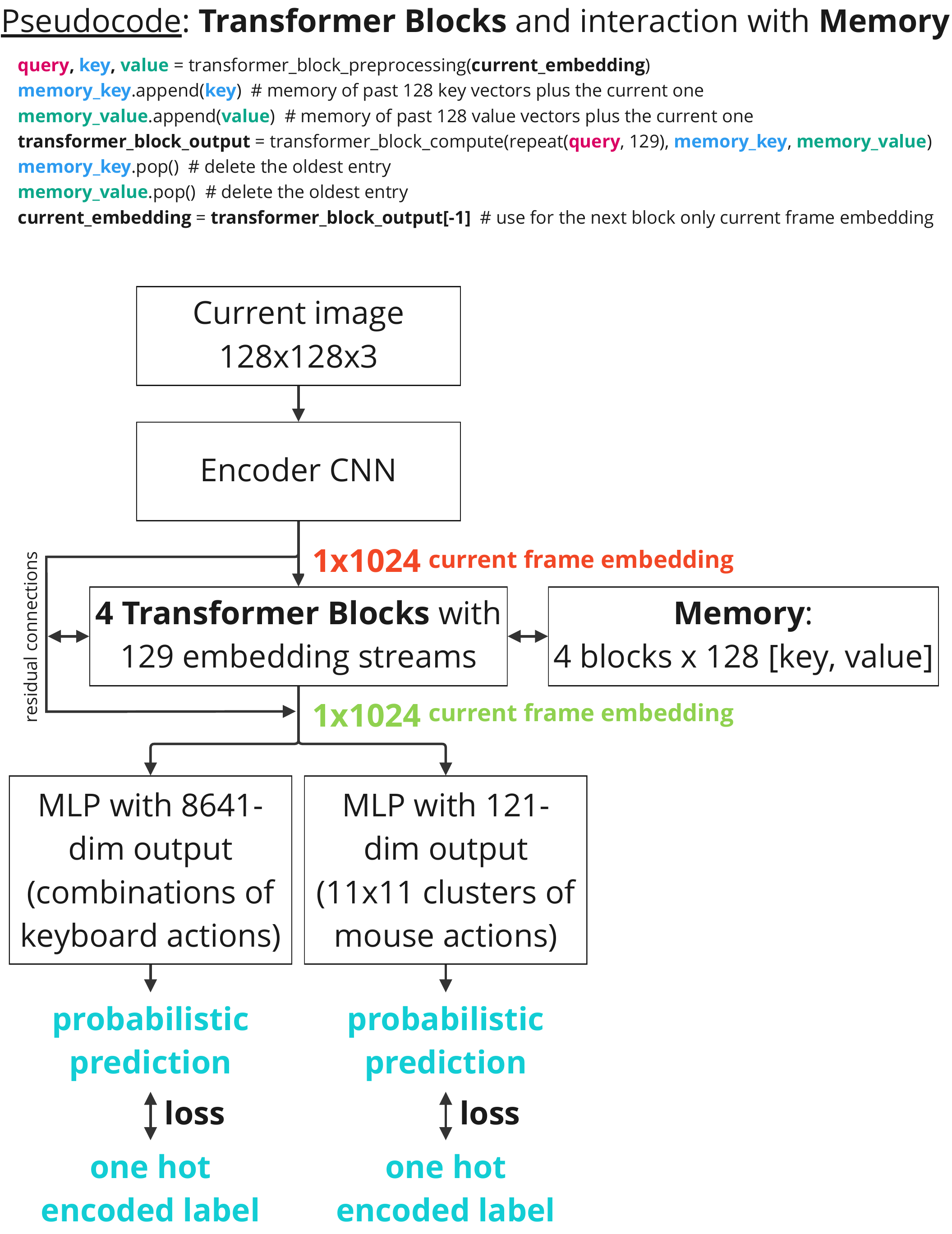}
    \caption{VPT architecture.}
    \label{fig:vpt_architecture}
    \vspace{-0.2in}
\end{figure}

Our approach uses a provided VPT model \cite{baker2022video} for encoding a \textit{situation} in a latent space (see Figure \ref{fig:vpt_architecture}). The model uses the IMPALA \cite{EspeholtKavukcuoglu2018} convolutional neural network (CNN) as backbone for the encoding of individual images. The CNN network encodes each image into a 1024-dimensional vector. The stack of 129 CNN outputs passes through four transformer blocks (see Figure \ref{fig:vpt_architecture}). Additionally to the current frame, a memory stack stores the last 128 embeddings for each transformer block. The output of the last transformer block are 129 embedding vectors, each 1024-dimensional. The architecture discards 128 output embedding vectors of the last transformer block and processes further only the current's frame embedding vector. Two MLP output heads take as an input the current's frame embedding vector to predict actions. The first output head predicts a discrete action (one out of 8641 possible combinations of compound keyboard actions). The second output head predicts a computer mouse control as a discrete cluster of one of the possible 121=11x11 mouse displacement regions (±5 regions for X times ±5 regions for Y).
The architecture is shown in Figure \ref{fig:vpt_architecture}.

\subsubsection{Search-based BC}

Search-based behavioral cloning (BC) aims to reproduce an expert's behavior with high fidelity by copying its solutions from past experience. We define a \textit{situation} as a set $\{(o_{\tau}, a_{\tau})\}_{\tau=t}^{t+\Delta t}$ of consecutive observation-action pairs coming from a set of provided expert's trajectories, where ${\Delta t}$ is less or equal to the number of input slots of a transformer block that processes embedding vectors of input images.

We encode the expert's past \textit{situations} through a provided VPT model \cite{baker2022video}. Thus, we obtain a latent space populated by N-dimensional \textit{situation} points. Due to the expert's optimality assumption, we can assume that each \textit{situation} has been addressed and solved in an \textit{optimal} way. 

We encode each sampled \textit{situation} with the same network. Then, we search the nearest embedding point in the dataset of \textit{situation} points. Once the reference \textit{situation} has been selected, we copy its corresponding actions. After each time-step we update the current and reference \textit{situations}, by updating the queue of embedding vectors of images for the current \textit{situation}, while shifting to the next time-step in the recorded trajectory from the dataset for the reference \textit{situation}. To assess the similarity, we compute the L1 distance between the current \textit{situation} and the reference \textit{situation}. In most cases, the reference and the current \textit{situations} will evolve differently over time, thus, their L1 distance will diverge. Therefore, at each timestep we recompute the similarity of the current and reference \textit{situations}. A new search is performed whenever either one of two conditions is met:
\begin{itemize}
    \item The L1 distance between current and reference \textit{situations} overcomes a threshold (see red lines in Figure \ref{fig:divergence});
    \item The trajectory from the dataset has been followed for more than 128 time-steps (see blue lines in Figure \ref{fig:divergence}).
\end{itemize}

Choosing feature divergence as a criterion for controlling search comes with a major advantage: whenever the copied actions can not be performed (e.g. there are physical constraints that limit the agent movement space), the features will diverge even faster. Thus, our agent will quickly perform a new search and address the faulty \textit{situation}.

\begin{figure}[t!]
    \includegraphics[width=\columnwidth]{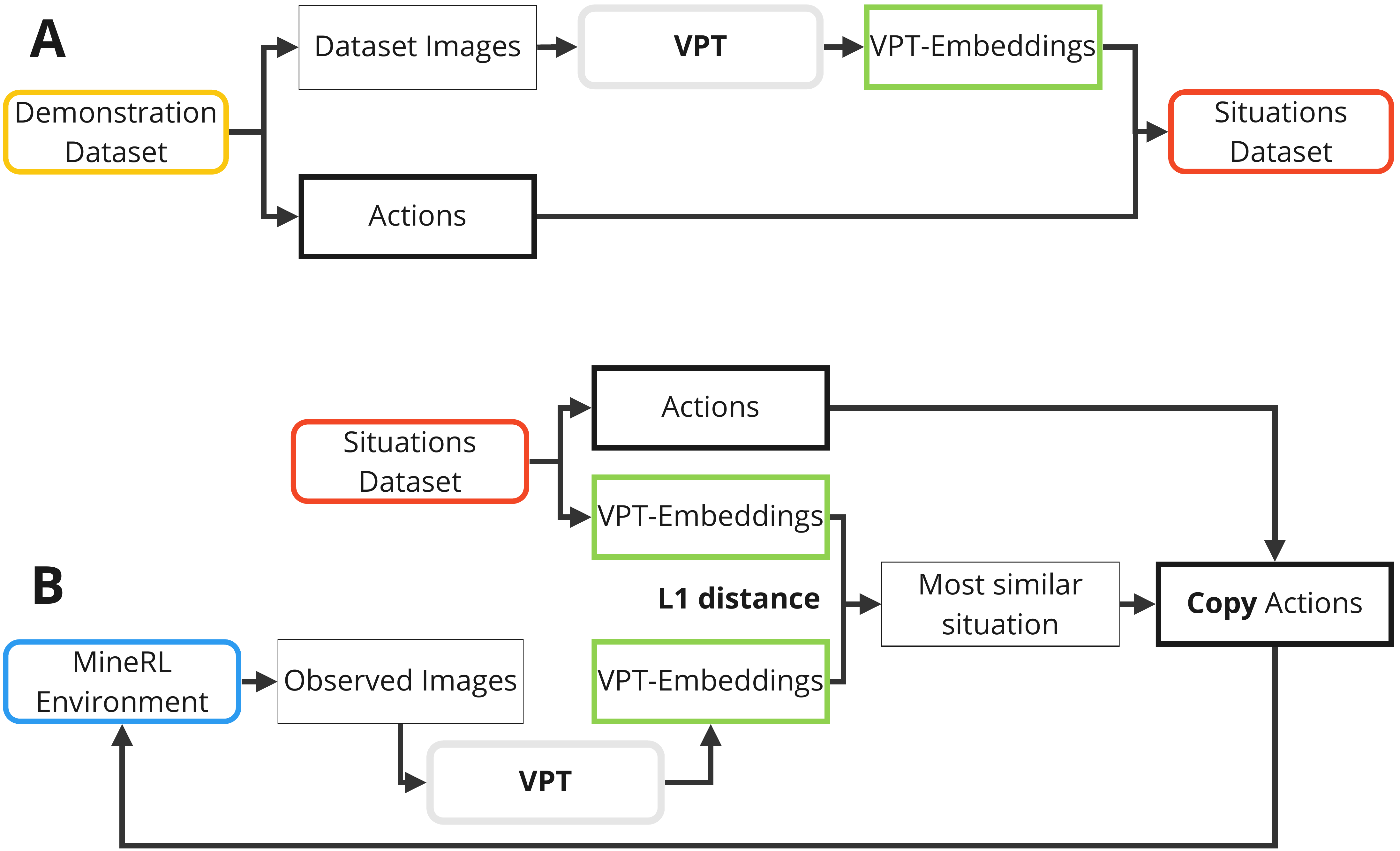}
    \caption{Our approach. (\textbf{A}) Training procedure. (\textbf{B}) Evaluation loop.}
    \label{fig:approach}
    \vspace{-0.2in}
\end{figure}

Our approach is illustrated in Figure \ref{fig:approach}. We refer to the generation of the latent space as the "training" procedure of our agent. Rather, it is a preprocessing step needed to ensure prior knowledge to our agent.

\section{Experiments and Results}

We applied our method to the MineRL BASALT Challenge 2022 \cite{ShahDragan2021}, where it ranked top of the leaderboard at the end of Round 1. The agent had to demonstrate human-like behavior while completing the tasks. Our agent produces visually human-resembling behaviour in the tasks.

\begin{table}[]
\addtolength{\tabcolsep}{-2pt}
\begin{tabular}{|p{2.6cm}||p{1cm}|p{1cm}||p{1.2cm}|p{1.2cm}|}
\hline
\textbf{Environment}  & \textbf{Spike Before} & \textbf{Spike After}  & \textbf{Window Before} & \textbf{Window \newline After} \\ \hline
FindCave &\tiny $.37\pm.02$ &\tiny $.17\pm.02$ &\tiny $.23\pm.05$ &\tiny $.15\pm.03$\\ \hline
VillageAnimalPen                               &\tiny $.37\pm.01$                              &\tiny $.16\pm.02$     &\tiny $.22\pm.04$ &\tiny $.15\pm.02$                               \\ \hline
MakeWaterfall                                        &\tiny $.37\pm.02$                               &\tiny $.16\pm.02$ &\tiny $.22\pm.04$ &\tiny $.15\pm.02$                                   \\ \hline
BuildVillageHouse                                    &\tiny $.37\pm.02$                               &\tiny $.18\pm.03$   &\tiny $.23\pm.04$ &\tiny $.16\pm.02$                                \\ \hline
\end{tabular}
\caption{Average L1 distance value and its standard deviation between current state and reference \textit{situation}. Spike denotes deviation based new search, Window denotes time based new search. Values shown are right before and after a new search.}
\label{quantitative}
\end{table}

In Table \ref{quantitative} we report quantitative measurements of the L1 distance before and after a new search for the best matching trajectory from the dataset. In all four tasks, we found that the average L1 distance after a search is much lower than before it.

A \textit{situation} encapsulates both current and past information. Therefore, at the very beginning of an episode, the \textit{situation} embedding may be not informative. To mitigate this, we allow the agent to \textit{warm up}, by keeping it still for the first second of a new episode. This way, the agent can gather some images and produce a more informative representation of the current \textit{situation}. Using the \textit{warm up} phase can be vital whenever the agent faces a dangerous \textit{situation} at the beginning of an episode, e.g. when spawning close to a lava pit.

\section{Discussion \& Conclusion}
Here we presented our approach that represents the control problem as a search problem over a latent space of partial trajectories (called \textit{situations}) from a dataset of experts' demonstrations. Our approach can effectively recover meaningful demonstration trajectories and show human-like behavior of an agent in the Minecraft environment. Possible directions for improving the approach are methods of self-supervised segmentation of important objects in first-person views \cite{melnik2021critic}, multi-modal fusion of segmented representations \cite{korthals2019jointly}, modularization of control \cite{melnik2019modularization}\cite{konen2019biologically} and involvement of working memory \cite{melnik2018world}.

\bibliography{bibliography.bib}
\bibliographystyle{ieeetr}
\end{document}